\documentclass[conference]{IEEEtran}
\IEEEoverridecommandlockouts
\usepackage{cite}
\usepackage{amsmath,amssymb,amsfonts}
\usepackage{graphicx}
\usepackage{textcomp}
\usepackage{xcolor}

\usepackage{booktabs} 
\usepackage{algorithm}  
\usepackage{algpseudocode}  
\usepackage{hyperref}

\def\BibTeX{{\rm B\kern-.05em{\sc i\kern-.025em b}\kern-.08em
    T\kern-.1667em\lower.7ex\hbox{E}\kern-.125emX}}
\begin{document}

\title{LLM With Tools: A Survey\\}

\author{\IEEEauthorblockN{1\textsuperscript{st} Zhuocheng Shen}
\IEEEauthorblockA{\textit{Department of Software Engineering} \\
\textit{Tongji University}\\
Shanghai, China \\
2150276@tongji.edu.cn}

}

\maketitle

\begin{abstract}
The integration of tools in augmenting large language models (LLMs) presents a novel approach toward enhancing the efficiency and accuracy of these models in handling specific, complex tasks. This paper delves into the methodology, challenges, and developments in the realm of teaching LLMs to use external tools, thereby pushing the boundaries of their capabilities beyond pre-existing knowledge bases. We introduce a standardized paradigm for tool integration guided by a series of functions that map user instructions to actionable plans and their execution, emphasizing the significance of understanding user intent, tool selection, and dynamic plan adjustment. Our exploration reveals the various challenges encountered, such as tool invocation timing, selection accuracy, and the need for robust reasoning processes. In addressing these challenges, we investigate techniques within the context of fine-tuning and in-context learning paradigms, highlighting innovative approaches to ensure diversity, augment datasets, and improve generalization. Furthermore, we investigate a perspective on enabling LLMs to not only utilize but also autonomously create tools, which may redefine their role from mere tool users to tool creators. Finally, we reproduced Chameleon's results on ScienceQA and analyzed the code structure.
\end{abstract}


\section{Introduction}
With the rapid development of artificial intelligence technology, large language models have become an indispensable part of our lives. The large models represented by ChatGPT\cite{chatgpt} and GPT-4\cite{achiam2023gpt} of OpenAI, with their excellent natural language understanding and reasoning abilities, can perform diverse tasks, greatly enriching the boundaries of human-computer interaction. However, these large language models still face many challenges in their application in specific fields.\par
Although large models have shown strong capabilities in a wide range of tasks, their performance is often unsatisfactory in certain professional fields, especially in scenarios that require precision\cite{lu2022survey} and real-time performance\cite{komeili2021internet}. This is mainly because large models are essentially probabilistic models, and their training data often cannot cover all real-time, specific information. Therefore, when dealing with problems involving complex mathematical operations or the latest knowledge, large models may experience hallucinations, where the output content or answers deviate from the actual situation.
In order to overcome this challenge, researchers have begun to explore new solutions. From the history of human development, we can draw inspiration that tools are the crystallization of human wisdom and an important extension of human abilities. By using tools, we can more effectively complete tasks, improve productivity and efficiency. This idea also applies to large language models. By endowing large models with the ability to use tools, we can make them more accurate and efficient in handling specific tasks.\par
Specifically, the use of tools for teaching large models needs to follow a holistic framework.\cite{qin2023tool} This framework starts with user instructions and requires the model to develop and execute executable plans related to the tool. To achieve this goal, the model first needs to have the ability to understand user intentions. This requires the model to accurately capture the real needs behind user instructions, thereby ensuring the pertinence and effectiveness of subsequent plans.\par
Secondly, the model needs to understand the functionality and usage methods of the tool. This includes understanding the basic operations, applicable scenarios, and limitations of the tool. By deeply understanding the tools, the model can more accurately determine when, where, and how to use them, thereby ensuring the smooth completion of tasks.\par
In addition, the model also needs to have the ability to decompose complex tasks into subtasks. By breaking down tasks into smaller and easier to handle units, the model can adjust plans more flexibly to adapt to different situations and needs. At the same time, the model also needs to have reasoning ability to dynamically adjust the plan during task execution, ensuring the smooth completion of the task.\par
By integrating the use of large models and auxiliary tools, we can not only improve the accuracy of task execution, but also to some extent expand the functional boundaries of large models, making them better serve humans. Research in this field will not only contribute to the development of artificial intelligence technology, but also have profound impacts on the progress of human society.

\section{Challenges}

\subsection{Time to invoke the tool}
It is crucial to correctly identify when to call tools during the use of large models. External tools should be called when the model itself cannot directly provide accurate answers or solutions, such as in scenarios that require access to real-time data, performing complex calculations within specific domains, or processing special format documents. At the same time, if information queries or processing can be derived from the existing knowledge and logical reasoning of the model, calling external tools may not be necessary, as this may lead to reduced efficiency, increased costs, and even erroneous results.\cite{yang2024gpt4tools,qiao2023making} In short, tools should not be called for the sake of calling them.

\subsection{Tool selection and accuracy}
Choosing the appropriate tools is crucial for ensuring the successful completion of tasks. However, as the number of available tools increases, ensuring the accuracy of each call becomes more difficult. Each tool has its specific advantages and limitations, and there may be performance differences between different tools. In complex reasoning processes, especially when involving non-linear reasoning or multi tool links, maintaining call accuracy becomes more challenging, as a small error can lead to the failure of the entire reasoning process.\cite{qin2023toolllm}

\subsection{Method of tool call}
Proper use of a tool is not only about knowing its existence, but also about understanding how to effectively call it. This involves understanding the tool interface, including the number, type, and value of incoming parameters. Incorrect parameters may cause the call to fail or return incorrect results. Therefore, understanding the API of each tool and how they respond to different types of requests has become very important.\cite{li2023api,xu2023tool,zhuang2024toolqa} So, how to efficiently operate these tools is crucial for avoiding call errors and optimizing usage processes.

\subsection{Robustness of reasoning process}
Throughout the entire reasoning process, the accumulation of errors may lead to the amplification of problems, thereby affecting the quality of the results.\cite{chen2023chatcot,song2023restgpt} Therefore, establishing a robust mechanism to detect and correct errors has become crucial. However, this process is full of challenges, especially in complex inference chains or multi-step operations. How to effectively backtrack and correct errors without causing the entire process to crash or be inefficient is a major challenge in the use of large models.

\subsection{Time efficiency}
Time efficiency is a key consideration in the use of large models. As the reasoning and tool invocation process becomes more complex, the time cost of the entire pipeline also increases. In addition, some tools may have delays, especially when it comes to network requests or processing large amounts of data. Therefore, how to optimize time efficiency and improve parallelism while ensuring the quality of task completion has become an important challenge when using large models.

\subsection{Generalization ability}
The implementation of General Artificial Intelligence (AGI) is one of the ultimate goals in the field of artificial intelligence. However, whether different tools can be called according to different scenarios to solve complex reasoning problems is still an open question. At present, technologies such as large language models are rapidly developing in this direction, but achieving complete AGI still faces many challenges. One of the key factors determining whether AGI can be achieved is whether the model has sufficient generalization ability to create or innovate based on existing tools for unfamiliar tools.

\section{Research Status}
The current research trend has shifted towards teaching large models how to properly utilize tools, while recent studies have begun to explore allowing large models to develop their own tools to more effectively solve problems.\cite{qian2023creator} In order to thoroughly understand the various stages of using or developing tools for large models, we introduce a standardized method for using external tools for large models. This paradigm provides a macro perspective that allows us to gain insight into the actual use of external tools in large-scale models.

\subsection{Paradigm of tool use}

Given a set of user instructions $U$, a toolset $T$, and an environment state space $E$, we can define:

\begin{itemize}
    \item A function $f_{\text{intent}}: U \rightarrow I$ that maps a user instruction to a user intent.
    \item A function $f_{\text{plan}}: I \times T \rightarrow P$ that generates a tool usage plan based on the user intent and the available toolset.
    \item A function $f_{\text{exec}}: P \times E \rightarrow E$ that executes a plan and updates the environment state.
    \item A function $f_{\text{feedback}}: E \rightarrow R$ that creates feedback results from the environment state.
    \item A function $f_{\text{perceive}}: R \rightarrow S$ that processes feedback and generates a summary.
    \item A function $f_{\text{adjust}}: S \times P \rightarrow P$ that adjusts the plan based on the summary.
\end{itemize}

The tool usage process can be modeled as follows:

\begin{enumerate}
    \item User issues an instruction $u \in U$.
    \item System identifies user intent $i = f_{\text{intent}}(u)$.
    \item System generates a plan $p = f_{\text{plan}}(i, T)$.
    \item System executes the plan in environment $e \in E$, resulting in a new state $e' = f_{\text{exec}}(p, e)$.
    \item System generates feedback $r = f_{\text{feedback}}(e')$.
    \item System processes feedback and generates a summary $s = f_{\text{perceive}}(r)$.
    \item System adjusts the plan based on the summary $p' = f_{\text{adjust}}(s, p)$.
\end{enumerate}

This process iterates until the task is completed.\par
\begin{figure}[t]
\centerline{\includegraphics[scale=0.7]{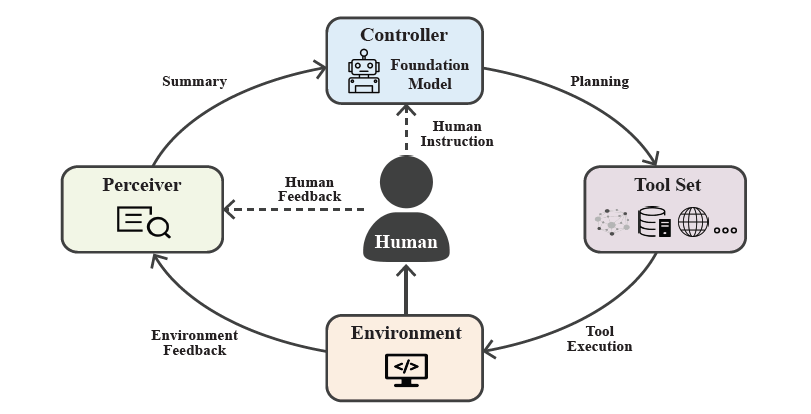}}
\caption{Whole process of LLM using tools.}
\label{fig1}
\end{figure}

The entire process can be seen in Fig. \ref{fig1}.\par

In the exploration of utilizing external tools within large-scale models, a fundamental step involves the selection of an appropriate tool for a given task. This process is predicated on three prevalent approaches: Firstly, injecting knowledge into the large model through the method of fine-tuning. This approach enables the model to assimilate specific expertise related to the tool and its application. Secondly, leveraging the in-context learning capabilities of the model, which allows it to infer how to utilize the tool based on provided examples or hints within its current context. Lastly, an innovative strategy involves enabling the model itself to generate tools tailored for specific tasks. Each of these strategies presents a unique pathway towards enhancing the model's functionality and its proficiency in executing tasks that necessitate external tools. These methodologies will be delineated further in the subsequent sections of this paper, offering a comprehensive analysis of their mechanisms and implications in the context of large-scale model utilization.

\subsection{Fine tuning}
To enhance our understanding of fine-tuning practices, it is imperative to devise a special form of dataset, denoted as $C^*$. For the purposes of demystifying the fine-tuning procedure and providing a foundational concept of the dataset structure, our approach includes a simplification of both the dataset and the fine-tuning process.\cite{schick2024toolformer} The aforementioned dataset is comprised of a sequence of samples ranging from $x_1$ to $x_j$. Each sample adheres to the following configuration:
\begin{equation}
X^* = w_{1:i-1}; E(c_i; R_i); w_{i:n} \label{eq1}
\end{equation}
\begin{equation}
E(c;r) = \textless TOOL\textgreater\; a_{c}(i_{c})\; !\;  r\; \textless/TOOL\textgreater \label{eq2}
\end{equation}

Herein, $E(c;r)$ represents a unique syntactical construct employed to delineate the invocation of tools. Within this context, \textless TOOL\textgreater and \textless /TOOL\textgreater  are utilized as boundary markers signalling the initiation and conclusion of a tool invocation, respectively. Meanwhile, "!" functions as a delineator, distinguishing between the invocation command and the resulting outcome of the tool's operation. Concretely, $a_{c}$ denotes the designated tool name for invocation, $i_{c}$ refers to the required input for said tool, and $r$ encapsulates the end result post-tool execution.

It is noteworthy that $w_{1:i-1}$ and $w_{i:n}$ are not inherently part of the tool invocation; rather, they are a sequence of words generated by a large-scale model drawing upon its extensive internal knowledge base. These words, in conjunction with the details of the tool invocation, collectively constitute an exhaustive sample.

Utilizing dataset $C^*$, one can engage in training a comprehensive, unaltered large-scale model, $M$. Throughout the fine-tuning phase, the language is employed in a standardized format to structure the objective function. In the course of assimilating knowledge from $C^*$, the model is incrementally equipped to make informed decisions regarding the timing, placement, and methodology of specific tool utilization, influenced by the context and internal feedback mechanisms.

In particular instances, as the model navigates an input sequence and identifies a marker akin to \textless TOOL\textgreater, it recognizes the necessity to summon an external tool. Subsequently, the model deciphers the name, input, and anticipated outcome associated with the tool, formulating a request directed towards the external tools accordingly. Following the receipt of the execution results from the tool, the model tactfully integrates this data into the continuing output sequence, thereby fulfilling the task of text generation to completion. The overall process of the inference stage can refer to Algorithm 1.

\begin{algorithm}[t] 
\label{algo1}
\caption{Text Generation with Tool Integration}     
\begin{algorithmic}[1]      
\Procedure{InferWithTool}{$model, input\_nl, maxlen$}      
    \State $output \gets []$  
    \State $i \gets 0$  
    \While{$i < maxlen$ \textbf{and} not $EndOfText$($output$)}      
        \State $predicted\_token \gets$ $model$.predict($input\_nl$)  
        \State $output$.append($predicted\_token$)  
        \State $i \gets i + 1$  
          
        \If{$predicted\_token = \langle TOOL \rangle$}      
            \State $tool\_query \gets$ extract\_query($output$) \Comment{Extract subsequent tokens after $\langle TOOL \rangle$}  
            \State $tool\_response \gets$ call\_tool($tool\_query$) \Comment{Invoke relevant tool}  
            \State $output$.append($\langle TOOL \rangle$)  
            \State $output$.append($tool\_response$)  
            \State $output$.append($\langle /TOOL \rangle$)  
            \State $i \gets i +$ length($tool\_response$) $+ 2$ \Comment{+2 for $\langle TOOL \rangle$ and $\langle /TOOL \rangle$}  
        \EndIf  
    \EndWhile  
      
    \State $final\_text \gets$ extract\_tool\_responses($output$) \Comment{Extract tool responses enclosed by $\langle TOOL \rangle$ and $\langle /TOOL \rangle$}  
    \State \textbf{return} $final\_text$  
\EndProcedure  
\end{algorithmic}  
\end{algorithm}

\subsection{Fine tuned dataset construction}
To enhance the precision and professional integrity of the content concerning the exploration of fine-tuning technology, we must first elucidate its foundational principle: the adaptive calibration of models employing annotated datasets to elevate their efficacy in designated tasks or instrumental applications. A primary impediment in this endeavor is securing high-caliber datasets for fine-tuning purposes. This manuscript endeavors to examine various methodologies for dataset acquisition and to expound upon their merits and drawbacks comprehensively, thereby providing efficacious strategies for the refinement of extensive models.
\par
An initial strategy is to derive datasets from empirical human tool interaction, as trailblazing frameworks such as WebGPT\cite{nakano2021webgpt} and WebCPM\cite{qin2023webcpm} have exemplified. The dataset thus procured boasts considerable strengths: exceptional data integrity, congruity with actual human tool engagement patterns, and relevance. Notwithstanding these benefits, the drawbacks of this method are conspicuous: the financial onus associated with labor and the restrictive nature of data aggregation owing to the dependence on specific tools or their combinations, impeding the encapsulation of diverse operational milieus.
\par
To surmount these constraints, scholars have embarked on employing advanced models as instructive archetypes and assembling proprietary datasets via the in-context learning ability of Large Language Models (LLMs). Emblematic implementations of this approach include constructs such as Toolformer\cite{schick2024toolformer} and ToolCoder\cite{zhang2023toolcoder}, which emulate human tool interaction sequences to synthesize ample datasets for fine-tuning. This paradigm mitigates labor expenses and encompasses an expansive spectrum of scenarios and tool amalgamations.
\par
Nonetheless, the potential for data contamination or distortional bias in the datasets curated by LLM cannot be disregarded, as it may deleteriously affect the fine-tuning outcomes of substantial models using instruments. Within the context of dataset compilation, an unwavering emphasis on data quality is paramount, supplemented by the institution of robust measures to augment the caliber of the data.
\par
In the corpus of extant scholarly work, some academics have delved into this predicament with a critical lens and have articulated potential resolutions.

\subsubsection{Diversity}
In examining the fine-tuning of large models, a key challenge is to ensure that the applications derived are both highly original and varied. While traditional in-context learning methods have helped large-scale models to learn and simulate related tool applications, the issue of generating numerous overlapping or similar applications cannot be overlooked. To address this, research by entities such as GPT4Tools\cite{yang2024gpt4tools}, ToolAlpaca\cite{tang2023toolalpaca}, and ApiBank\cite{li2023api} has underscored the vital role that data variety plays. Often, these studies employ ablation analyses to vividly showcase the deep influence that a diverse dataset has on enhancing the model's applicational competencies post fine-tuning when tested. Findings indicate that datasets featuring a broad spectrum of case types substantially boost the model's ability to generalize tool applications.\par

\begin{figure}[t]
\centerline{\includegraphics{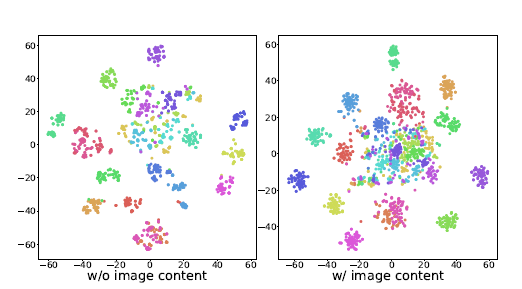}}
\caption{Diversity of w/o and w/ image content.}
\label{fig2}
\end{figure}

One especially innovative approach within the GPT4Tools\cite{yang2024gpt4tools} research leverages multimodal data. This tactic aims to merge tips pertaining to images with tools, and then harnesses Chatgpt\cite{chatgpt} among other linguistic models to produce content that's intricately linked to the imagery. This approach does more than just guarantee content relevance—it also stretches the breadth of instructional diversity by tapping into the inherent variety within images. The impact of using multimodal data and not using multimodal data on data diversity can be seen from Fig.\ref{fig2}\par

Additionally, LLMs in action: Tool learning through simulated trial and error\cite{wang2024llms} adopts long short-term memory principles to pioneer a data assembly method that mirrors the trial and error dynamics between LLMs and tools. In this model, the short-term memory attends to the latest experimental record, meticulously chronicling each interaction and its outcome so that the model can swiftly learn from fresh successes and missteps, modifying its approach on the fly. Conversely, the long-term memory archives a comprehensive log of trial and error experiences over time, aiding the LLM in developing deep insights into tool application. With each new experiment, the short-term memory gets an immediate update to reflect the latest interactions. Meanwhile, the long-term memory feeds into the LLM's context, furnishing essential historical knowledge to prevent the re-creation of previously attempted tool applications. This unique memory framework empowers the LLM to evolve and gather experience constantly, significantly diversifying case generation.\par

\subsubsection{Data augmentation}
In the realm of real-world problem solving, the employment of tools necessitates intricate reasoning sequences and numerous tool engagements, leading to what is identified as the context sample issue. While traditional rudimentary tools utilize datasets to educate large-scale models to some degree, the models' capacity to generalize remains underdeveloped for intricate challenges demanding combinatory logic and multiple tool interactions. Despite the large model's capacity for emergence, its efficacy is hampered in scenarios involving complex multi-tool interactions.\par

It is imperative to note the existing research’s scant consideration in assessing the generalization capacity of models on complex tool usage test sets, following the refinement of large models with simple tool usage datasets. However, the concept of Skills-in-Context prompting\cite{chen2023skills} sheds critical light: in the domain of in-context learning, furnishing compound examples within prompts markedly enhances the model’s aptitude to manage complex issues that involve multiple tool engagements.\par

To more adeptly simulate and prepare large-scale models for complex multi-tool interaction scenarios, ToolAlpaca\cite{tang2023toolalpaca} leverages a multi-agent framework, this strategy engenders exhaustive and varied tool usage examples by mimicking the multi-stage tool usage dynamic in reality. Utilizing three distinct agents - the user agent, assistant agent, and tool executor agent, this approach emulates the real-world tool utilization workflow akin to the React\cite{yao2022react} framework operation.\par

The user agent, portraying the tool’s end-user, is tasked with concocting tool usage directives and addressing inquiries from the assistant based on the ongoing interaction context. It guarantees task directive diversity and richness via various prompt templates. The assistant agent, embodying a tool-savvy assistant, assimilates directives from the user agent, stipulates subsequent maneuvers, elects suitable tools and functionalities, and fabricates the ultimate response. The tool executor agent is charged with mimicking the tool’s operational process, fielding requests from the assistant agent, and formulating responses leveraging the tool’s functionalities. This methodology streamlines the intricacy of actual API invocations through simulation and generation capabilities and has been verified for accuracy and efficacy.\par

The collaborative dynamic among these agents unfolds as follows: initially, the User agent formulates instructions informed by tool data; subsequently, the Assistant agent selects fitting actions and their inputs, pending the tool executor agent’s simulated execution and feedback; this sequence perpetuates until the assistant agent deems it has garnered sufficient information to address the user's directives. Through this multi-agent interplay, real tool usage scenarios are emulated, generating comprehensive and varied tool usage exemplars.\par

This methodology's merit lies in addressing the issue of manually drafting multi-stage tool usage instances, considerably reducing labor costs. Moreover, it aligns more closely with the multi-stage dialogue and the experimental approach prevalent in real-world tool usage, augmenting the trained model’s practicality and generalization capability.\par

Beyond this, TaskBench\cite{shen2023taskbench} and ToolLLM\cite{qin2023toolllm} have proposed innovative algorithms for orchestrating multiple rounds of tool interactions. TaskBench\cite{shen2023taskbench} interlinks tools based on dependencies (resource or temporal), transfigures the toolbox into a tool graph, and from it derives diverse subgraphs to inversely steer the creation of user directives, task sequences, and tool invocation graphs for large models. ToolLLM\cite{qin2023toolllm} employs RapidAPI’s levels to randomly select tools from identical categories or sets, deriving APIs from these tools to formulate directives. This ensures the plausible coverage of a tool cluster, showcasing significant diversity and practical value. These investigations not only augment the dataset of multi-tool invocation scenarios but offer more expansive and effective training methodologies for large models in processing complex challenges.\par

\subsection{Other issues during the fine-tune phase}
When examining the deployment of Large Language Models (LLMs) within the domain of tool learning, a number of critical challenges emerge. Foremost, the discussion on LLMs' capability to assimilate new tools through simulated trial and error encounters a principal issue: ensuring the model's continuous evolution and adaptation to new tools whilst preserving the skills previously acquired. LLMs in the Imaginarium: tool learning through simulated trial and error\cite{wang2024llms} highlights that traditional fine-tuning techniques frequently result in catastrophic forgetting. To mitigate this, experience replay has been identified as an efficacious approach when incorporating new tools.\par

Nonetheless, the expandability of LLMs as the repertoire of tools broadens warrants attention. Specifically, initial methodologies, such as representing each tool with a distinct token and learning its embedding as suggested by Toolkengpt\cite{hao2024toolkengpt} and similar models, offer convenience. Yet, as the tool inventory expands, the effectiveness of such approaches may diminish due to the increased model complexity and potential for performance decline. The challenge, therefore, lies in devising a strategy that not only handles a vast array of tools efficiently but also sustains model performance.\par

Moreover, several studies focus merely on the model's ability to invoke the correct tool API by name, overlooking critical evaluation metrics like the accuracy, number, and variety of input parameters.\cite{zhuang2024toolqa} In practical scenarios, these parameters are crucial for the correct functioning of tool applications. Consequently, future research should aim for a more comprehensive and in-depth assessment of model performance upon the integration of new tools.\par

Lastly, the potential of LLMs to refine their capabilities based on user feedback is an area worth exploring. In real-world settings, user feedback is a pivotal factor for ongoing model refinement. Models such as WebGPT\cite{nakano2021webgpt} utilize reinforcement learning techniques to adjust and optimize outputs based on user inputs, enhancing model adaptability and user satisfaction. Integrating user feedback into model training processes is thus imperative for elevating LLM performance in tool learning applications.\par

\subsection{In context learning without retrieval}
The advancement of in-context learning in large language model (LLM) utilization represents significant strides in intelligent systems' capabilities to comprehend and deploy external tools. Initially, we examine the in-context learning phase involving select tools. Here, the model leverages tool instruction documents or sample tasks as prompts, engaging in few-shot learning. Notable examples of such models include Hugginggpt\cite{shen2024hugginggpt} and Chameleon\cite{lu2024chameleon}. These approaches are esteemed for their plug-and-play functionality, diminishing the learning costs and time required for users to interact with the model.\par

As model complexities and task intricacies escalate, Hugginggpt\cite{shen2024hugginggpt} underscores a heightened demand for extensive and high-caliber task demonstrations, deeming them crucial for in-context learning effectiveness. This raises the inquiry: Is there a definitive writing methodology or a template to guide the composition of effective task demonstrations? This inquiry pertains not solely to the form and substance of the prompts but also to the integration of divergent tools and methodologies.\par

\begin{figure}[t]
\centerline{\includegraphics{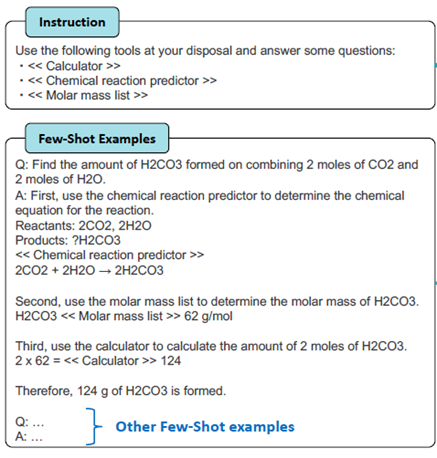}}
\caption{Example of MultiTool-CoT prompt.}
\label{fig3}
\end{figure}

To address this issue, the MultiTool-CoT\cite{inaba2023multitool} method introduces a novel tactic. This technique utilizes chain-of-thought (CoT) prompting\cite{wei2022chain}, enabling LLMs to invoke sequential external tools judiciously within appropriate reasoning phases, adhering to a few-shot learning paradigm. The essence of this methodology is that by demonstrating to the model the dexterous utilization of various tools during distinct problem-solving junctures, both the precision and the efficiency of task resolution can be markedly uplifted. Specifically, this approach not only fortifies the LLM's versatility and adaptability but also broadens its scope of application, thereby enhancing its overall problem-solving prowess.\par

The enactment of the MultiTool-CoT\cite{inaba2023multitool} framework hinges on a profound comprehension of the logical progression in task processing. Through meticulously formulated prompts, the model is steered to pinpoint and execute pivotal procedural steps, discerning the optimization of timing and methodology for incorporating external tools. The merit of this tactic is its tailor-made processing of intricate tasks, paired with a proficient exploitation of multi-tool collaborative potential. Fig.\ref{fig3} is an example of the MultiTool-CoT\cite{inaba2023multitool} framework.\par

Nevertheless, this approach unveils specific constraints, most notably a confined assortment of tools, often tailored exclusively for particular domains or datasets. Such restrictions can impede the model's generalizability and scalability, potentially hampering its applicability and effectiveness across a spectrum of diverse tasks.\par

\subsection{In context learning with retrieval}
There are indeed several challenges associated with the integration and utilization of a multitude of tools. The foremost concern is the limitation concerning the length of context. As the assortment of tools expands, it becomes unavoidable for the context to grow verbose, impeding the process of effectively procuring essential information. In addition, there is the question of whether language models possess the capability to accurately strategize and select the pertinent tools amidst a plethora. In reaction to these predicaments, researchers have proposed in context learning with retrieval."\par

What, then, does retrieval entail? Here, retrieval is defined as the search conducted based on specific tasks or tools. For instance, Chatcot\cite{chen2023chatcot} utilizes the tasks retrieved as exemplars within context learning to aid in the decomposition of tasks by language models (LLMs), such as TaskMatrix.AI\cite{liang2023taskmatrix}, which emphasizes tool retrieval itself, permitting LLMs to initially devise a plan, thereafter retrieve specific tools through search, and subsequently integrate prompts to ensure the model aptly invokes the suitable tools.\par

Current scholarly discourse introduces two distinct methodologies for retrieval:
The first methodology advocates for modularity, exemplified by TaskMatrix.AI\cite{liang2023taskmatrix} and Restgpt\cite{song2023restgpt}, which categorize tools, initially searching for appropriate categories, followed by the LLM selecting specific tools for invocation.
Alternatively, the dense retrieval model operates through the adjustment of embedding vectors, as seen in TaskMatrix.AI\cite{liang2023taskmatrix}, Chatcot\cite{chen2023chatcot}, and Gorilla\cite{patil2023gorilla}, among others. These adopt text embedding to find the response that semantically resonates with the query.\par

Moreover, Tool Documentation Enabling Zero Shot Tool Usage with Large Language Models\cite{hsieh2023tool} indicates that in scenarios involving the usage of tools on a large scale, marrying retrieval mechanisms with tool documentation can surpass performance solely dependent on demonstrations. Concurrently, Gorilla's inquiry underscores the pivotal role of the retriever's selection on the outcomes.\par

What benefits does this approach offer? Primarily, it accommodates the employment of an extensive array of APIs, with research from ToolLLM\cite{qin2023toolllm} evidencing commendable transferability of this method. Secondly, through the retrieval of API documentation, LLMs' context learning faculties can be wholly harnessed to accurately invoke APIs. This approach not only circumvents errors, for instance, inaccuracies in parameter numbers and other hallucinations but also enables the content of API documentation to be updated as necessary, swiftly adapting to increments.\par

Regarding the API documentation itself, a review of extant research suggests that an efficacious document should encompass the following components:
\begin{itemize}
\item API Name : The API name provides an abstract of the API. The name should be clear and precise in natural language and avoid ambiguity with other API names.
\item Parameter List : The parameter list for an API includes the input parameters and return value, and each parameter has a parameter name, parameter description, data type, and default value.
\item API Description : The API description contains more information about what the API does, how it works, what are its inputs and outputs, and any potential errors or exceptions that may be raised.
\item Usage Example : Providing usage examples for the Api.
\end{itemize}
\par

\subsection{Other issues during the in context learning phase}
In the discussion of in context learning, we confront a significant challenge that merits attention. When a large language model (LLM) is preliminarily configured with predefined action paths, such configurations often display inherent static qualities, making it challenging for these models to flexibly adapt to evolving feedback from their environments. This rigidity can result in delayed responses to unforeseen events and, in some cases, may lead to failure due to a single oversight. Thus, it's crucial to explore strategies to mitigate this limitation.\par

\begin{figure}[t]
\centerline{\includegraphics{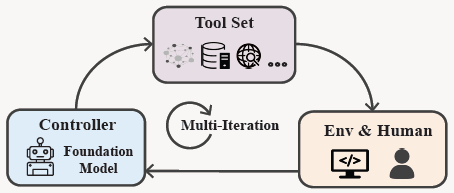}}
\caption{Process of online planning.}
\label{fig4}
\end{figure}

A notable example of overcoming this hurdle is seen in the implementation of an online planning mechanism by Chatcot\cite{chen2023chatcot}. In this model, every decision is considered within the broader context of the entire dialogue, rather than as an isolated event. This approach allows the model to adjust its future actions in real-time, basing these adjustments on the outcomes of preceding decisions and external feedback. Through online planning, the model achieves a level of flexibility that allows it to adapt to dynamic dialogue environments effectively. Importantly, despite the adoption of online planning, the efficiency of Chatcot remains intact, as evidenced by the minimal increase in token consumption. Fig.\ref{fig4} shows the process of online planning.\cite{chen2023chatcot}\par

Another strategy, paralleling the design seen in Assistgpt\cite{gao2023assistgpt}, involves the introduction of a learner module to address execution errors. When the model encounters errors during task execution, the Learner module steps in to identify and attempt to rectify these mistakes. This self-corrective mechanism enhances the model's resilience and adaptability, facilitating more robust performance.\par

Lastly, the use of a DFSDT algorithm for tree search, as employed by ToolLLM\cite{qin2023toolllm}, represents a third strategic approach. The DFSDT algorithm allows models to explore decision trees more thoroughly, considering multiple possibilities at each layer. This method empowers the model to systematically evaluate various prospective solutions and select the most suitable one when faced with complex challenges. By incorporating DFSDT, the model benefits from improved decision-making quality and accuracy in complex situations.

\subsection{LLM creates its own tools}
In the previous discussion, we mainly focused on how to utilize existing tools to assist large language models (LLMs) in task processing. However, recent research has proposed an innovative paradigm of allowing large models to autonomously create and use tools, which to some extent goes beyond the traditional role of tool users. Specifically, research represented by Creator\cite{qian2023creator} and Large language models as tool makers\cite{cai2023large} has demonstrated the potential of LLM for tool generation and correct solutions on unknown problems.\par

The core of the Creator\cite{qian2023creator} framework consists of four closely connected modules. Firstly, The creation stage guides LLM to generate the required tools through fixed demonstrations. This stage focuses on stimulating the creativity of LLM and generating auxiliary tools that match the problem. Next, The decision stage utilizes the demonstration method again to guide LLM in generating code or methods on how to use these newly created tools. This stage fully demonstrates LLM's concrete reasoning ability, which is the ability to make logical reasoning based on specific situations. Subsequently, The execution stage is responsible for executing these created tools, which requires LLM to have the ability to solve specific problems and transform abstract tools into practical operations. Finally, The recognition stage makes necessary modifications and optimizations to tools or decisions to ensure that they can more effectively serve the target task.\par

The Creator\cite{qian2023creator} framework also proves that tools created by large language models have certain transferability in similar scenarios. However, a detailed answer has not yet been provided on how to effectively reuse these tools. On this issue, Large language models as tool makers\cite{cai2023large} proposes a solution to achieve tool reuse by introducing a dispatcher module. The dispatcher module searches and filters out suitable tools from all tool caches generated in the past. However, the module uses a naive context learning method without retrieval function when selecting tools. This means that as the number of tools continues to increase, The context length may become a significant challenge as it directly relates to the length of text that the model can handle. Therefore, when the number of tool caches surges, the accuracy of reusing tools solely based on context learning and whether context length limitations can be guaranteed have become a question worthy of in-depth exploration. At present, there is a lack of systematic research in this field. However, we can draw on the retrieve concept mentioned earlier to seek a solution. We can build a retrieval system that contains rich tool information. When reusing tools is needed, the retrieval system can quickly locate the appropriate tool and provide it to the dispatcher module for selection. This solution not only helps to reduce the limitation of context length, but may also improve the accuracy and efficiency of tool reuse.

\subsection{Other methods}
In the process of exploring automated tool calls to answer specific questions, an effective solution is to select the most appropriate tool to call by comprehensively evaluating the similarity between the question and the potential answer, as well as the description of each tool. Taking Gear\cite{lu2023gear} as an example, this scheme shows an efficient strategy to integrate semantic and schema similarity.
First of all, Gear deeply analyzes the internal relationship between the problem and each tool description through the semantic similarity scoring mechanism. The core of this step is to use natural language processing technology, such as word embedding model or deep learning algorithm, to quantify the semantic proximity between the problem text and tool description. The calculated semantic similarity score, Gear is able to initially screen out toolsets that are highly relevant to the topic of the problem.\par
Then, Gear further introduces the concept of pattern similarity scoring. At this stage, the system will generate one or more simple preliminary answers according to the nature of the question, and compare these answers with the output generated by each tool. The purpose of pattern similarity scoring is to evaluate whether the output of the tool is consistent with the preliminary answer in structure, content or format. By comparing the answers with the output of the tool, Gear can screen out tools that can produce answers that meet expectations.\par
Finally, Gear combines semantic similarity score with pattern similarity score to give each tool a comprehensive score. This scoring mechanism comprehensively considers the semantic relevance between tools and questions and the accuracy of tool output, so as to ensure that the selected tools can meet the needs of problem solving to the greatest extent. Based on these scores, Gear can automatically select the most appropriate tool to call and generate the corresponding answer.
However, it is worth noting that although gear, a tool selection scheme based on similarity scoring, has high efficiency and accuracy in theory, it still faces some challenges in practical application. First, because this method needs to call all tools at once to obtain the output, it may incur high computational cost in the face of large-scale data sets. Secondly, the solution is currently mainly limited to the solution of a single problem, and may not be flexible enough for complex scenarios that need to deal with multiple problems. In addition, for some specific areas or professional issues, the existing toolset may not be fully covered, which will also limit the scope of application of this method.\par
Therefore, future research can further explore how to optimize the tool selection algorithm to reduce the computational cost, and expand its scope of application to support more complex problem-solving scenarios.

\section{Reflection and Outlook}
Although there is currently a lot of research on the use of tools for large models, there are still many research directions in this field. In response to the limitations of current research in this field and future research directions, I propose the following in-depth reflections.
\subsection{Research on tool scheduling topology optimization and graph decision making}
How to enable LLM to derive the optimal tool scheduling topology that can accurately generate answers and responses given a set of predefined tools and user queries is a worthwhile research direction. In addition, due to the fact that tool calls and task decomposition are essentially a graph data structure, it is worth studying how to establish bridges between graph data structures, graph neural networks, and large models for auxiliary decomposition and decision-making.

\subsection{Time optimization of model pipelines}
Although sequential, multi-step methods are a common practice for handling complex queries, they may not be optimal or most suitable for practical use. In some cases, using nested, parallel, or iterative function calls may be more effective. How to design algorithms that can shorten the entire pipeline time without reducing accuracy is also a worthwhile research direction.

\subsection{Model optimization and continuous learning}
In order to adapt to a large number of constantly increasing and evolving tools, large models require continuous learning and optimization. How to achieve plug and play without compromising model performance is an important research direction.

\subsection{Compensate for API errors and prevent error cascading}
API calls to other models or tools may result in incorrect results, and it is also possible that the initial plan was not properly planned. So, LLMs should learn to evaluate the reliability of APIs and summarize and recover from errors, avoiding step-by-step errors is also a worthwhile research direction.

\subsection{Learn to use tools}
How to best teach a large model how to use tools is still an unresolved issue. The accuracy of existing research in finding the most suitable tool is far from satisfactory. Therefore, finding the most relevant tools is also a research direction.

\subsection{Pretrain tool\_augmented LLMs}
Existing research is mostly based on fine-tuning pre trained models or tool calls for in context learning. Due to the diversity of tool call cases, we can try to pretrain tools to enhance LLMs.

\section{Experiments on Chameleon}
\subsection{Experiment settings}
To access the various GPT models securely, we utilized the Azure OpenAI Service, a trusted and compliant cloud platform. In this experiment, we use GPT-3.5 (gpt-35-turbo-16k-0613). We test on 4241 test examples on ScienceQA by using method CoT and Chameleon.

\subsection{Experiement results}
\begin{figure}[t]
\centerline{\includegraphics{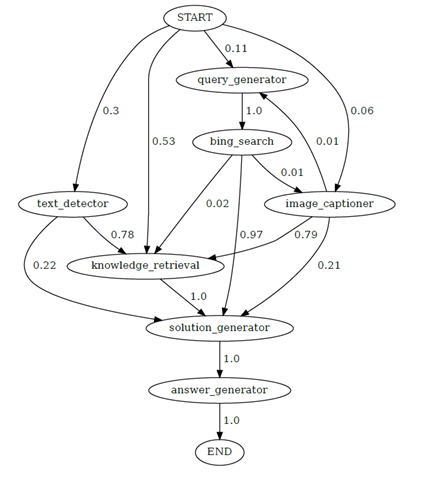}}
\caption{Chameleon's state transition diagram.}
\label{fig5}
\end{figure}

\begin{figure}[t]
\centerline{\includegraphics{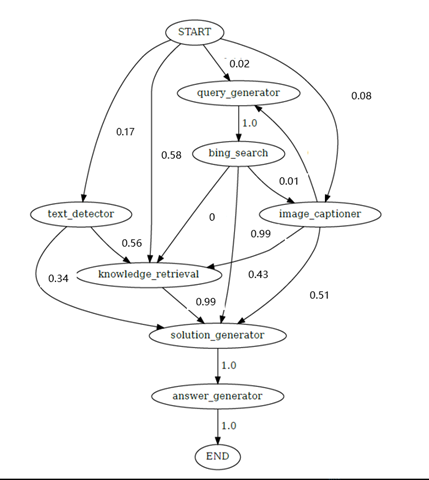}}
\caption{Reproduced state transition diagram.}
\label{fig6}
\end{figure}

\begin{figure}[t]
\centerline{\includegraphics[scale=0.6]{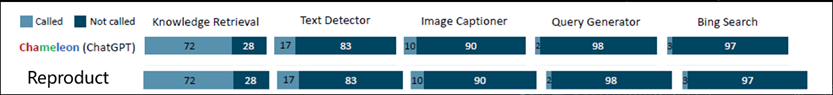}}
\caption{Call scale diagram.}
\label{fig7}
\end{figure}

The QA accuracy can be seen from Table\ref{table1} and Table\ref{table2}, and the reproduced results are basically similar to the results given in the article, with only a decimal point difference. For Chameleon's state transition diagram and the reproduced state diagram, which can been seen in Fig.\ref{fig5} and Fig.\ref{fig6}, are slightly different. The call scale diagram, which can be seen in Fig.\ref{fig7}, is completely the same.

\subsection{Code structure and overall process of Chameleon}
Chameleon\cite{lu2024chameleon} first defines a class solver that contains various modules, such as image\_capterer, bing\_search, solution\_generator, etc. For the overall process, we propose a pseudocode, which is shown in Algorithm 2.\par

Below is a specific explanation of the role of each module.
\begin{itemize}
    \item load\_data : loading data
    \item get\_question\_text : obtaining problem text
    \item predict\_modules : predicting the modules to be executed
    \item build\_prompt for sg\_chameleon : constructing prompt texts in chameleon mode
    \item build\_prompt for sg\_cot : constructing prompt texts in cot mode
    \item image\_capterer : processing images
    \item text\_detector : processing text in images
    \item bing\_search : calling search engines
    \item solution\_generator : generating answers based on predicted results
    \item answer\_generator : generating final answers
\end{itemize}

\begin{algorithm}[t]
\caption{Main Execution Process}  
\label{algo2}
\begin{algorithmic}[1]  
\State \textbf{Input:} Parse input parameters (\texttt{parse.args})  
\State \textbf{Create:} Solver object  
\For{\textbf{each} iteration}  
    \State Initialize problem cache  
    \State Load problem instance  
    \State Predict execution module  
    \For{\textbf{each} execution in module}  
        \State Execute MODULE (input, output)  
    \EndFor  
    \State Compare predictions with answers  
    \State Update statistics  
    \State Save results  
\EndFor  
\State End of loop  
\State Summarize and save the final statistical data  
\end{algorithmic}  
\end{algorithm}

\begin{table}[t]  
\label{table1}
\centering 
\caption{The QA accuracy of the paper and its reproduction}
\begin{tabular}{lccccc} 
\toprule 
Model & Tuned\_Params & ALL & NAT & SOC & LAN \\  
\midrule  
CoT\_Paper       & 0M  & 78.31 & 78.82 & 70.98 & 83.18 \\  
CoT\_Us           & 0M  & 78.38 & 80.11 & 69.52 & 82.00 \\  
Chameleon\_Paper  & 0M  & 79.93 & 81.62 & 70.64 & 84.00 \\  
Chameleon\_Us     & 0M  & 79.56 & 81.71 & 69.29 & 83.45 \\  
\bottomrule 
\end{tabular}  
\end{table}  
  
\begin{table}[t]  
\label{table2}
\centering  
\caption{The QA accuracy of the paper and its reproduction}
\begin{tabular}{lccccc}  
\toprule  
Model & TXT & IMG & NO & G1-6 & G7-12 \\  
\midrule  
CoT\_Paper       & 77.37 & 67.92 & 86.13 & 80.72 & 74.03 \\  
CoT\_Us           & 77.13 & 68.07 & 85.85 & 81.38 & 72.84 \\  
Chameleon\_Paper  & 79.77 & 70.80 & 86.62 & 81.86 & 76.53 \\  
Chameleon\_Us     & 79.72 & 71.10 & 85.99 & 81.97 & 75.21 \\  
\bottomrule  
\end{tabular}  
\end{table}

\bibliographystyle{unsrt} 
\bibliography{refs}

\end{document}